\documentclass{article}
\usepackage{iclr2021_conference,times}
\usepackage{hyperref}
\usepackage{url}
\usepackage{graphicx}
\usepackage{listings}
\title{Tailoring Generative AI Chatbots for Multiethnic Communities in Disaster Preparedness Communication: Extending the CASA Paradigm}

\author{Xinyan Zhao\\
Hussman School of Journalism and Media\\
University of North Carolina-Chapel Hill\\
\And
Yuan Sun\\
College of Journalism and Communications\\
University of Florida\\
\And
Wenlin Liu\\
College of Journalism and Communications\\
University of Florida\\
\And
Chau-Wai Wong\\
Electrical and Computer Engineering\\
NC State University}

\iclrfinalcopy 
\begin{document}

\maketitle

\begin{abstract}
This study is among the first to develop different prototypes of generative artificial intelligence~(GenAI) chatbots powered by GPT-4 to communicate hurricane preparedness information to diverse residents. Drawing from the Computers Are Social Actors paradigm and the literature on disaster vulnerability and cultural tailoring, we conducted a between-subjects experiment with 441 Black, Hispanic, and Caucasian residents of Florida. Our results suggest that GenAI chatbots varying in tone formality and cultural tailoring significantly influence perceptions of their friendliness and credibility, which, in turn, relate to hurricane preparedness outcomes. These results highlight the potential of using GenAI chatbots to improve diverse communities’ disaster preparedness.
\vspace{2mm}
\end{abstract}

\section*{Lay Summary}
This study looks at how generative artificial intelligence~(GenAI) chatbots help individuals from different racial groups prepare for hurricanes in Florida. We create and test GPT-4 chatbots that use different tones and tailor their messages to fit people’s cultural backgrounds. Our results show that chatbots with an informal tone are seen as more friendly, whereas those with either a formal tone or one that fits people’s cultural backgrounds are seen as more credible. Credible chatbots encourage people to seek, share, and act on disaster information. Overall, GenAI chatbots can create human-like and culturally-sensitive disaster communication for diverse communities.

\section{Introduction}
Generative artificial intelligence~(GenAI), an emerging form of AI that produces customized content through generative models, presents transformative potential in disaster preparedness, response, and recovery (Bari et al., 2023). The past decade has seen a rapid development of AI-powered technologies used for disaster management. In disaster preparedness, AI has enhanced disaster predictions and early warnings during earthquakes, hurricanes, and floods (Lin et al., 2021). In disaster rescue and relief, AI supports crowd-sourced applications such as ``live maps'' that harness collective efforts to improve disaster relief outcomes (Ghaffarian et al., 2019).

Among various AI applications, chatbots stand out as one of the most promising tools for disaster management agencies to communicate disaster preparedness information to the public. Still in its infancy, research has demonstrated the versatility of chatbots in improving customer service experience, healthcare outcomes, and organization–public relationships (e.g., Men et al., 2023). In disaster and risk communication, generative AI chatbots have the potential to transform one-way, generic communication into interactive and tailored information for different community members. Such cultural tailoring is crucial for serving culturally and linguistically diverse communities, which often become hard-to-reach populations due to language barriers and ingrained communication preferences (Howard et al., 2017; Liu \& Zhao, 2023). These communities are also the focus of disaster vulnerability research, as they suffer from deteriorated disaster outcomes more than other communities (Jay et al., 2023).

Recent human-computer interaction (HCI) research has demonstrated that advanced machine-learning technologies have significantly improved chatbots’ ability to mimic human interactions~(Goyal et al., 2018) and made communication more interactive and personalized (Hancock et al., 2020). However, distrust remains an issue due to perceived deficiencies of chatbots in empathy and contextual understanding (e.g., Zhou et al., 2023). In high-stakes contexts like disaster communication, the use of chatbots could pose significant challenges, especially for multiethnic communities. This is due to historical factors such as cultural insensitivity and systemic racism, which contribute to long-standing mistrust among marginalized communities toward the government (Best et al., 2021).

In disasters, individuals’ interaction with public agencies is primarily driven by their informational needs (Liu, 2022; Liu \& Zhao, 2023). To help organizations meet the informational needs of diverse cultural groups, this study designs and tests GenAI chatbots as interactive, human-like information provision agents, grounded in the computers are social actors (CASA) paradigm (Nass \& Moon, 2000), through the lens of conversational tone (Kelleher \& Miller, 2006) and cultural tailoring (Kreuter \& McClure, 2004). The strategic communication literature on conversational tone highlights informal communication style (e.g., Men et al., 2023), which can improve public understanding and acceptance of complex, technical disaster information. Cultural tailoring, integrating both surface and deeper cultural elements (Resnicow et al.,1999) such as language, names, values, and beliefs, can enhance the chatbot’s relatability and credibility, thereby fostering user engagement. By creating GPT-4 chatbots that vary in tone and cultural tailoring, our study is among the first scholarly attempts to investigate how diverse communities perceive and interact with GenAI chatbots, and how this new approach can improve their disaster preparedness outcomes.

In a between-subjects experiment with 441 Black, Hispanic, and Caucasian Florida residents, participants interacted with chatbots powered by OpenAI’s GPT-4 API for hurricane preparedness, varying by tone formality and cultural tailoring, followed by a questionnaire. We then analyzed how the variations in chatbot communication influenced participants’ chatbot perceptions, and subsequently, their hurricane preparedness outcomes.

\subsection{Theorizing GenAI Chatbots in Disaster Communication: The CASA Paradigm}

Chatbots are software applications that engage users through natural language processing (NLP) capabilities (Shawar \& Atwell, 2007). With recent advances in machine learning, chatbots have evolved significantly in mimicking human-like attributes (Goyal et al., 2018). Hancock et al. (2020) argued that AI-mediated communication could transform interpersonal communication by tailoring responses to individual user preferences and communication styles. However, distrust, or algorithm aversion (Dietvorst et al., 2015), stemming from a perceived lack of human-like qualities such as empathy or contextual understanding (Zhou et al., 2023), remains a challenge. Longoni et al. (2019) found that users tend to distrust AI health systems due to the perception that algorithms cannot account for unique individual differences the way a human might.

To address these challenges, the CASA paradigm provides a foundational framework for understanding how humans interact with technology. According to the CASA paradigm, people often respond to computers and other media as if they were actual social beings, applying social rules and behaviors to these interactions on a subconscious level (Nass et al., 1994; Nass \& Moon, 2000). Extending this theory to media, Lombard and Xu (2021) further argued that various cues from new technologies can evoke a sense of social presence, making these technologies appear more “alive” or “social.” Thus, fostering humanness may be critical to overcoming distrust and promoting user engagement, especially in high-stake contexts such as disaster communication. For instance, Go and Sundar (2019) found that using highly anthropomorphic visual cues, such as human-like profile pictures, can compensate for the negative effects of a machine-like conversation style.

Building upon the CASA paradigm in the context of disaster communication, this study explores two key anthropomorphic cues—conversational tone and cultural tailoring—to develop chatbots as proxies for information sources, specifically serving as AI agents to deliver disaster information. A conversational tone is defined as a “conversational human voice” and “an engaging and natural style of organizational communication” (Kelleher, 2009, p. 177). According to the CASA paradigm, by adopting a conversational tone, a crucial anthropomorphic attribute in interpersonal communication, chatbots can better mimic human communication to create a friendly and sociable presence and trigger social responses from users, making interactions feel more natural and engaging. Compared to traditional one-way broadcasting communication, a conversational tone within the organizational context can better facilitate open dialogue and bring about positive relational outcomes such as relational satisfaction (Kelleher \& Miller, 2006). Indeed, Men et al. (2022) found that chatbots with a conversational tone can improve perceptions of organizational listening and transparency, leading to more positive organization–public relationships.

Cultural tailoring emerges as the second critical dimension, essential for meeting the diverse information needs of multiethnic communities. While prior studies have explored basic personalization techniques grounded in the CASA paradigm, such as using names (e.g., Chen et al., 2021) or demographic-based message tailoring (e.g., Whang et al., 2022), a deeper understanding of cultural differences in AI interactions remains lacking. Based on the CASA paradigm, when chatbots are culturally tailored to match the user’s racial identity, users may subconsciously respond as if they are interacting with an in-group member. This tailored experience could enhance user trust and their willingness to engage deeply with the chatbot. This effect might be more pronounced in disaster scenarios, where certain racial communities exhibit a heightened risk perception. Cultural and socioeconomic differences can affect user interactions with chatbots and disaster preparedness outcomes (Appleby-Arnold et al., 2018), highlighting the need for more research on culturally tailored chatbots and their effectiveness in disaster preparedness.

Taken together, enhancing the human-like qualities of GenAI chatbots through conversational tone and culturally tailored communication may help improve disaster communication outcomes for multiethnic communities.

\subsection{Disaster Preparedness Outcomes}
As disaster preparedness is central to effective disaster management, the current study focuses on three important indicators of disaster preparedness as identified in the literature (e.g., McLennan et al., 2020; McNeill et al., 2013): disaster-related information seeking intention, information sharing intention, and disaster preparedness efficacy.

Disaster information seeking is an active communication action in problem-solving processes (Kim et al., 2010). Information seeking is defined as the intentional scanning of the environment to gather information on a particular topic (Zhao \& Tsang, 2021). In disaster communication, information seeking is vital in sensemaking processes. When faced with crises, threats, or other uncertain situations, individuals are naturally prompted to seek information as a coping mechanism to mitigate tensions and anxiety. However, research also suggests that disasters may overwhelm individuals, leading to avoidance rather than active information seeking (Seeger et al., 2003). Promoting information-seeking behavior is therefore needed to engage communities in disaster preparedness. With the rise of misinformation during disasters (Hunt et al., 2020), promoting information seeking from official sources, such as disaster management agencies, is particularly urgent.

Meanwhile, disaster information sharing intention refers to the willingness to transmit relevant information to a larger crowd to facilitate disaster coping or problem-solving. As Kim et al. (2010) suggest, a problem ``is easier to solve when it comes to a problem of a group rather than that of an isolated individual'' (p. 149). Active information sharing thus helps improve individuals’ and communities’ disaster preparedness.

Finally, a sense of disaster preparedness captures the perceived confidence and efficacy in preparing for and coping with a disaster. Research has proposed multiple dimensions under this construct, including a cognitive dimension that taps into individuals’ risk awareness, disaster knowledge, and physical preparation intentions, as well as a psychological dimension that indicates individuals’ psychological readiness for disasters (Paton, 2003).

Having identified the primary outcomes of disaster preparedness, the following sections detail communication tone and cultural tailoring, two key mechanisms in the disaster vulnerability and HCI literature.

\subsection{The Role of Communication Tone in Chatbot Disaster Communication}
The type of human--computer interaction within disaster communication is unique. While both information and social support needs arise from disasters, individuals’ interaction with public agencies is more likely driven by one’s informational needs. This tendency is evidenced by individuals’ strategic construction of disaster communication ecologies, where certain information sources---such as media and public agencies---are predominantly used for obtaining disaster information, while interpersonal communication becomes the preferred source for social support (Liu, 2022; Liu \& Zhao, 2023; Zhao \& Liu, 2024).

Information delivery and provision thus become the primary function in the organizational use of technology for disaster communication. To that end, research has identified key mechanisms---from both the communicator and recipient’s sides---that can facilitate public acceptance of disaster risk information. From the delivery side, an informal communication style has been particularly noted in the literature. Due to the knowledge gap and the technical nature of disaster information, an accessible, translational communication style can improve public understanding of science and risk information (e.g., Mabon, 2020). This echoes the rich body of strategic communication research on conversational human voice (e.g., Men et al., 2023), characterized by informal communication style. This style features casual, colloquial language, in contrast to a formal style that uses authoritative and structured language (Gretry et al., 2017).

From the message recipients’ side, we focus on two mediating factors of particular relevance to communicating disasters to multiethnic communities, friendliness and credibility perceptions. The decline of trust in public institutions has been particularly felt among marginalized, multiethnic communities (Best et al., 2021). Among racial minorities, community members are less likely to seek disaster information from public agencies than their White counterparts, likely due to factors such as systemic racism and hostile perceptions of government agencies (Best et al., 2021). The alienating experience of seeing public agencies as “others” rather than “one of us” can be greatly mitigated by the perception of friendliness. Just as intergroup contact theory confirms the critical effect of friendships on enhancing intergroup experience (e.g., Pettigrew, 1998), we similarly hypothesize that perceived friendliness can improve disaster communication outcomes.

Meanwhile, as rumors and misinformation often emerge from a disaster, a strong credibility perception towards official sources would motivate the public to seek information from public agencies over other sources (Austin et al., 2012; Zhao \& Tsang, 2021). Credibility perception, therefore, serves as an equally important mediating mechanism of interest.

Organizations can strategically use an informal communication style to reduce the social distance from the public and make the interaction personal to foster engagement and trusting relationships~(Liu et al., 2020). HCI literature suggests this can be achieved through the human-like design of chatbots. Specifically, Chaves and Gerosa (2021) demonstrated that chatbots employing a more casual, friendly tone were perceived as more likable, leading to higher user satisfaction than those using a formal tone. Go and Sundar (2019) identify three factors that can increase the level of humanness of chatbots---anthropomorphic cues, human identity cues, and conversational cues---which, in turn, lead to more favorable attitudes and greater behavioral intentions to return to a given website.

However, existing studies have shown that an informal communication style can backfire when perceived as inappropriate in the organizational context (Gretry et al., 2017), particularly during disasters when the public expects the government to be authoritative and credible (Zhao \& Zhan, 2019). Taken together, the formality of a GenAI chatbot’s tone is hypothesized to impact user perceptions of the chatbot, specifically by decreasing perceived bot friendliness but increasing perceived bot credibility.

\textbf{H1:} The tone formality of a GenAI chatbot (a) negatively affects diverse residents’ perceptions of the chatbot’s friendliness and (b) positively affects credibility.

According to Johnson and Grayson (2005), perceived friendliness is crucial for establishing trust in user–chatbot relationships. This perception can evoke benevolent attitudes toward chatbots (Baek et al., 2022). The perceived friendliness can enhance the sense of social presence or the perception of interacting with a real human (Konya-Baumbach et al., 2023), and thus is more likely to increase trust and credibility. For example, a study revealed that the perceived friendliness of a chatbot, achieved through social and informal language cues, signaled its credibility and increased users’ willingness to disclose personal information (Jin \& Eastin, 2022). Based on the evidence, we propose the following hypothesis:

\textbf{H2:} The perceived friendliness of a GenAI chatbot is positively associated with its perceived credibility.

Further, we assess agencies’ use of GenAI chatbots to improve individuals’ disaster preparedness outcomes. Previous research has revealed that perceived friendliness positively affects users’ evaluation of chatbots (Wang et al., 2023). By eliciting benevolent feelings toward the bots, friendliness perception helps promote prosocial and other compliant behaviors (Baek et al., 2022). For example, Verhagen et al. (2014) found that perceived friendliness positively predicted user satisfaction. Recent research has also shown that chatbots can improve communities’ disaster-coping experience by promoting community members’ communicative activeness and collective efficacy (e.g., Tsai et al., 2019). To test whether the use of the GenAI chatbot can enhance disaster preparedness outcomes through perceived friendliness and credibility, we propose the following hypotheses:

\textbf{H3:} The perceived friendliness of a GenAI chatbot positively relates to disaster preparedness outcomes, including (a) information-seeking intention, (b) sharing intention, and (c) preparedness.

\textbf{H4:} The perceived credibility of a GenAI chatbot positively relates to disaster preparedness outcomes, including (a) information-seeking intention, (b) sharing intention, and (c) preparedness.

To test whether and how perceived friendliness and credibility perceptions may mediate the relationship between the GenAI chatbot’s tone and disaster preparedness outcomes, we propose the following research question:

\textbf{RQ1:} How does the tone formality of a GenAI chatbot influence diverse residents’ disaster preparedness outcomes (including information seeking, sharing, and preparedness) through chatbot perceptions (perceived friendliness and credibility)?

\subsection{The Role of Culturally Tailored GenAI Chatbots for Diverse Communities}
As mentioned, it is crucial for GenAI chatbots to adapt to the cultural nuances of multiethnic communities, as ethnic minorities with cultural and socioeconomic differences can exhibit different patterns of interactions with chatbots, producing diverging disaster preparedness outcomes (Appleby-Arnold et al., 2018). Culture, broadly defined, can include race, ethnicity, gender, and sexual orientation. Given our focus on multiethnic communities, this study focuses on race/ethnicity as the major indicator of culture. Cultural tailoring, which adapts messages to the cultural characteristics of a specific group, helps communication interventions effectively address a given group’s decision-making needs and interests (Kreuter \& McClure, 2004). As more disaster and risk communication research focuses on explaining and reducing disaster coping disparities among diverse communities, cultural tailoring has been proposed as a promising approach to bridge disparities (Huang \& Shen, 2016). Ma and Zhao (2024) posited that cultural tailoring involves highlighting specific risks and coping implications for distinct cultural groups. For example, by differentiating between marginalized communities (e.g., the Hispanic or Black community) and the general population, disaster communication can be designed to meet the unique information needs of diverse groups and reduce disparities.

The strategies of cultural tailoring can be categorized into surface tailoring and deep tailoring~(Resnicow et al., 1999). Surface-level communication tailoring adapts to a culture’s “superficial” aspects, including language, appearance, and diet. By contrast, deep tailoring engages with a community’s social, historical, and psychological characteristics, such as values, traditions, and norms—considered as the culture’s deep structure (e.g., Kreuter et al., 2004). A meta-analysis on cultural tailoring in cancer communication shows that culturally tailored cancer communication had a small significant effect on persuasion, and deep tailoring had a more substantial impact than surface tailoring (Huang \& Shen, 2016).

The concept of cultural tailoring has rarely been investigated in the literature of disaster communication and HCI. Contextualizing cultural tailoring in disaster communication and GenAI chatbots, this study examines both surface-level and deep-level tailoring for more inclusive chatbot-human interactions. Surface-level tailoring includes adapting the chatbot’s language, slang, and the use of culturally familiar names to resonate with specific cultural groups. Drawing on the CASA paradigm, this approach can be viewed as a form of anthropomorphism, where culturally relevant cues enhance the chatbot’s perceived humanness and relatability. In contrast, deep-level tailoring involves embedding deeper cultural elements---such as values, traditions, religious beliefs, historical contexts, and cultural narratives---into chatbot design. This deep-level approach remains underexplored within HCI, partially because existing theories like the cue effects of the HAII-TIME model (Sundar, 2020) primarily focus on surface-level cues and heuristics rather than examining how deeply embedded cultural values and beliefs can foster trust and engagement among diverse groups. By integrating both aspects of cultural tailoring into GenAI--human interactions, our study has the potential to enrich user engagement and disaster preparedness outcomes for multiethnic communities. Given the lack of research on culturally tailored GenAI chatbots in disaster preparedness, we propose the following research questions to explore their impact on diverse communities’ perceptions of the chatbots and subsequent disaster-related outcomes:

\textbf{RQ2:} How does a culturally tailored GenAI Chatbot influence diverse residents’ chatbot perceptions~(perceived friendliness and credibility)?

\textbf{RQ3:} How does a culturally tailored GenAI Chatbot influence diverse residents’ disaster preparedness outcomes (information-seeking intention, sharing intention, and preparedness) through chatbot perceptions (perceived friendliness and credibility)?

\section{Methods}
Following the approval of the University Institutional Review Board, we conducted an online between-subjects experiment from February to March 2024. Participants aged 18 and older in Florida were recruited through Prolific, an online research platform. Once enrolled, participants were directed to a Qualtrics survey and informed about using a GPT-4 chatbot, simulating a scenario of a local government’s hurricane preparedness education. Consented participants first completed screening and indicated their primary racial/ethnic identity. After reviewing the instructions, participants were randomly assigned via Qualtrics to one of the chatbot conditions programmed to generate unique uniform resource locators (URLs) for different conditions. Clicking on the assigned URLs redirected participants to our web browser interface to start their interactions with designated chatbots. Afterward, they were instructed to return to Qualtrics to complete the survey by answering questions measuring the outcome and demographic variables. At the end, participants were debriefed, compensated, and provided with links to emergency response resources from federal and state emergency response agencies.

\subsection{Participants}
We used quota sampling to represent Florida residents by race/ethnicity, intentionally oversampling minorities, including Blacks and Hispanics, to achieve an equal distribution among different racial/ethnic groups in our sample. This allowed us to understand chatbot utilization for hurricane preparedness among vulnerable communities. The final sample size was 441 after removing 11 responses that failed all attention check questions. A comparison of the demographics between the final sample and the excluded responses did not reveal any patterns associated with attention check failure. In the final sample, the average age was 38.38 (\textit{SD =} 14.08). The sample consisted of 148 (33.56\%) White/Caucasian, 150 (34.01\%) Black/African American, and 143 (32.43\%) Hispanic/Latino participants. Within the sample, 62.36\% were women, 34.47\% were men, and 3.17\% reported non-binary. Approximately 18.37\% had high school or lower education levels, 70.98\% had a partial or full college education, and 10.66\% had a graduate degree or above. The median household income was between \$25,000 to \$49,999.

\subsection{OpenAI’s GPT-4 API and Web Server}
Our chatbot utilizes OpenAI’s Chat Completions API, leveraging GenAI capabilities of an internally trained model designated as “gpt-4-1106-preview.” It is important to understand the distinction between GPT-4 API and ChatGPT. GPT-4 API provides a scalable and flexible framework that allows for programming-based customization of multiple chatbots, enabling their interactions with thousands of users simultaneously. This chatbot was developed from the open-source “openaichatexample” project on GitHub. We enhanced it by adding adaptive prompts for various experimental conditions, logging chat histories and timestamps for analysis, configuring the web server for load handling and pressure tests and managing daily data backup during deployment (for technical details, see Supplementary material P1). Our work selected GPT as the AI engine for our chatbot due to its superior performance compared to other large language models, as validated by extensive computing and NLP experiments (Duan et al., 2024; Ziems et al., 2024). The source code of our chatbot and its user instructions are available at: \url{https://bitbucket.org/leecwwong/ai_chatbot/}.

\subsection{GenAI Chatbot Manipulation} 
We developed different sets of system prompts as input for OpenAI’s GPT-4 API, aiming to train different versions of GenAI chatbots. Initially, we created a general prompt where GPT 4 was instructed to simulate an agent from a Florida emergency management team providing reliable information as a member of a government agency for participants across all conditions. We then created four specific sets of prompts as manipulations of the chatbot’s tone (formal vs. informal) and cultural tailoring (tailored vs. generic). Building on the literature (Gretry et al., 2017), the tone was trained to vary between an informal tone—characterized by casual language, acronyms, and emojis when appropriate—and a formal tone, characterized by official and authoritative language representing an official agency (for details, see Supplementary material P2).

Following the literature on cultural tailoring (Huang \& Shen, 2016), the culturally tailored chatbot adapted its conversation based on the participant’s race/ethnicity (a dynamic variable determined by their input). For example, this chatbot was prompted to “adapt your language, tone, slang, acronyms, emojis, and other textual cues as appropriate based on the \%race\%,” and “provide credible and accurate information, knowledge, and/or support that addresses the common needs and challenges faced by the \%race\% communities in Florida” (for details, see Supplementary material P2). With Hispanic and Black participants, it used a culturally familiar name, provided bilingual support (Hispanic only), suggested the inclusion of culturally relevant items in the emergency kit, acknowledged the unique needs and concerns of the community, and emphasized family unity and cultural preservation. In contrast, the generic chatbot provided hurricane preparation information without cultural tailoring. Note that cultural tailoring was not applied for White/Caucasian participants, who all received the generic condition, as disaster preparedness information predominantly addresses the needs and perspectives of this majority group, inherently aligning with their cultural background and serving as the baseline for comparison. Examples of chat scripts are included in Supplementary material P3.

\subsection{GenAI Chatbot-User Interaction Process}
Participants were invited to evaluate a prototype GenAI chatbot for disaster preparedness education. They were instructed to imagine that a hurricane was forecasted for the upcoming weeks and interact with the chatbot to learn more about hurricane preparedness. Participants were randomly assigned to chatbot conditions on Qualtrics through a two-step process. First, a random number generator determined whether they would interact with a chatbot using an informal or formal tone. Then, a Qualtrics randomizer function assigned participants to either a culturally tailored chatbot, matching their primary racial/ethnic identity, or a non-tailored control chatbot (for technical details, see Supplementary material P4). Supplementary Table S4.2 confirms a successful randomization check. A custom URL, incorporating these assignments, directed participants to the appropriate web browser-based interface for their interaction. The web server received the encoded assigned condition through the custom URL and transmitted the relevant prompts to OpenAI’s GPT-4 API at the backend. This ensured that participants experienced interactions aligned with their assigned conditions without seeing the actual prompts. Each participant’s message and the entire chat history are sent to GPT-4 for reasoning and response. If multiple messages are sent before a reply, GPT-4 addresses them collectively. Participants were asked to interact with the chatbot for at least five minutes, and our server logged their chat history and response times.

\subsection{Pilot Test}
A pilot test ($n = 40$) was conducted to ensure the web server’s functionality and the manipulation’s appropriateness. Participants from various racial communities responded favorably to the chatbot, supporting its appropriateness and utility. Their feedback led to adjustments in our prompts for the main study. These adjustments focused on prioritizing the provision of information over making promises about actions, such as dialing 911 or contacting hotlines, starting interactions in English before presenting options for additional languages, and changing language when requested. 

\subsection{Measurement}
\textbf{Bot Credibility.} We used five items from Flanagin and Metzger (2003) to measure bot credibility: “To the best of your judgment, is the chatbot [believable, accurate, trustworthy, biased, or complete]?” (1 = not at all, 7 = very much). The mean was 5.92 (\textit{SD =} 1.00, Cronbach’s alpha = .88). 

\textbf{Bot Friendliness.} We used ﬁve items from Koh and Sundar (2010) to measure bot friendliness: “To me, is the chatbot [empathetic, personal, warm, willing to listen, or open]?” (1 = not at all, 7 = very much). The mean was 5.30 (\textit{SD =} 1.31, alpha = .91).

\textbf{Information seeking intention.} Adapted from Zhao and Tsang’s (2021) scale, participants reported the likelihood they would seek hurricane preparation information from state government agencies and local government agencies on a 7-point scale ranging from 1 “not likely at all,” to 7 “extremely likely.” The mean of information seeking intention was 5.36 (\textit{SD = }1.02, alpha = .75).

\textbf{Information sharing intention.} Following Zhao and Tsang (2021), participants also indicated the extent to which they were likely to share the information provided with people they know, such as family, friends, and co-workers (\textit{M =} 5.54, \textit{SD =} 1.28, alpha = .80).	

\textbf{Disaster Preparedness.} Disaster preparedness measures were adapted from McNeil et al. (2013) and McLennan et al. (2020). Participants rated their level of agreement with six statements on a 7-point Likert scale. Example statements included “If a hurricane takes place, I know how to take proper actions to ensure my family’s and my safety” and “I feel confident about protecting my family and me against the negative impact of a hurricane.” (\textit{M =} 5.75, \textit{SD =} 0.85, alpha = .86). One item “I do not feel anxious even with an impending hurricane affecting my area” was removed due to its lower internal consistency in the sample.

\textbf{Covariate.} Disaster experience and racial identification were used as covariates. Disaster experience was measured through a binary variable indicating whether participants had previously experienced any hurricanes. To measure participants’ identification with their racial/ethnic group, we adopted three items from Leach et al. (2008). For instance, participants rated the statement “The fact that I am a [Hispanic/Latino] is an important part of my identity” on a 7-point Likert scale (\textit{M =} 4.88, \textit{SD =} 1.75, alpha = .90). The term within brackets corresponded to the participant’s self-reported racial/ethnic identity. 

\subsection{Analytical Scheme}
An exploratory computational analysis was first conducted on all chat scripts to identify the prevalent topics, providing the context for subsequent statistical analyses (for details, see Supplementary material P6). To test the hypotheses and research questions on chatbot perceptions and disaster outcomes, we conducted structural equation modeling through the R “Lavaan” package. Two models were analyzed, including the overall sample (\textit{N = }441) and a sub-sample comprising Hispanic and Black participants (\textit{n =} 293), which provided a more focused test of the effect of cultural tailoring. In the structural model, exogenous variables were the two manipulated chatbot conditions coded as binary variables and covariates (not shown in the figure for simplicity). Mediators included perceived cultural tailoring, bot friendliness, and bot credibility. Endogenous variables included information-seeking intention, information-sharing intention, and disaster preparedness (Figure 1). In the measurement model, latent constructs with fewer than three items were identified through all items. To identify latent constructs with more than three items, a parceling approach was used to create composite items (see notes of Figure 1 for details). Table 1 shows descriptive statistics and a correlation matrix of these variables for the full sample. Parameters were estimated by maximum likelihood. The model was evaluated using standard cutoff values for the model-data fit indices (Hu \& Bentler, 1999). The bootstrap method (\textit{N =} 5,000, biased corrected) was used to estimate indirect effects.

\begin{table}[!t]
\caption{Summary Statistics and Correlation Matrix}
\includegraphics[width=1.0\linewidth]{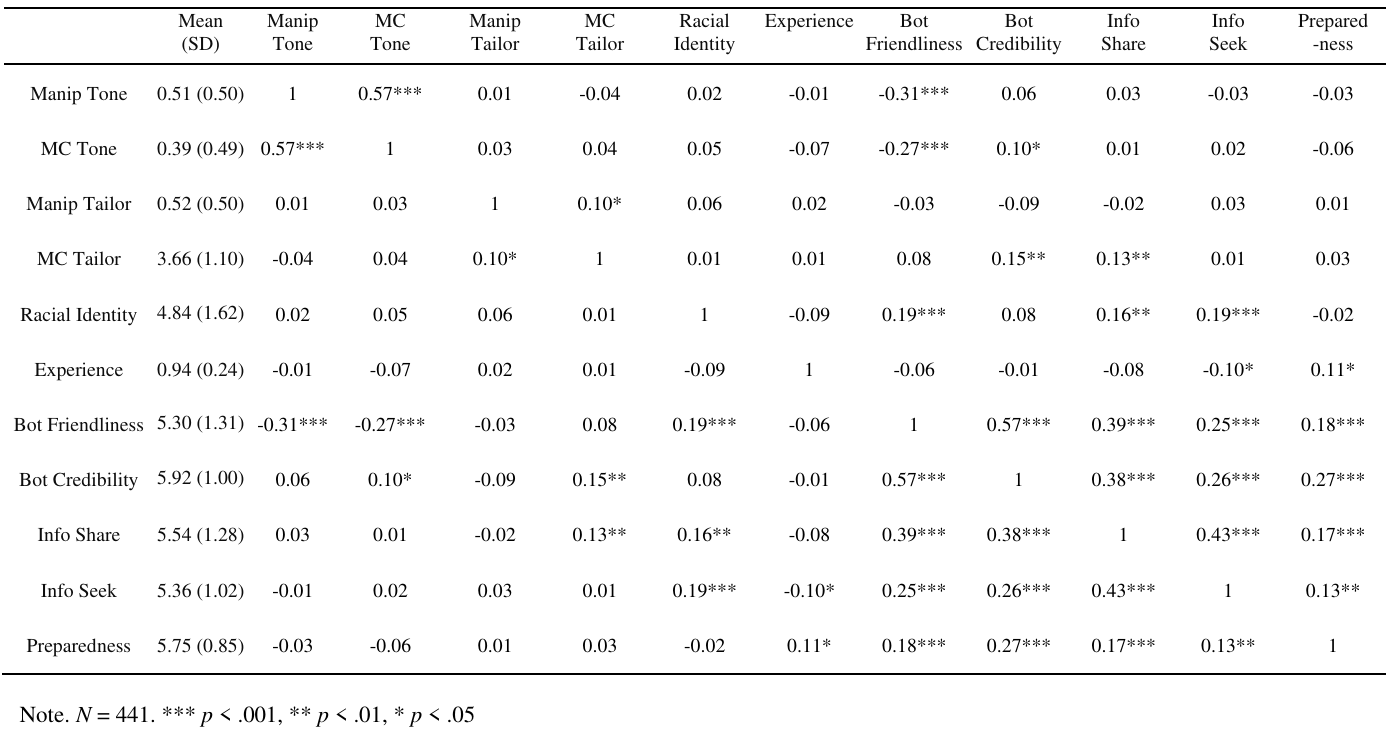}
\label{tab:corr}
\end{table}

\section{Results}
\subsection{Manipulation Checks}
On average, participants interacted with the chatbot for 6 min and 42 s (\textit{SD =} 4 min and 13 s). In total, participants inputted 3,615 textual entries, while the GenAI chatbots inputted 4,233 entries. Given potential randomness in texts generated by GenAI chatbots, we conducted two sets of manipulation checks testing both actual and perceived tone formality and cultural tailoring. First, we measured the actual levels of linguistic manipulation using a computational analysis of all chat scripts through OpenAI’s GPT-4 model (see Supplementary material P5 for the prompts). The actual level of cultural tailoring in the chatbot text (0--5) was measured by summing five binary indicators, such as a culturally familiar agent name or proposed language options (for details, see Supplementary material P5). The actual level of tone informality was measured based on the ratio of colloquial words, slang, acronyms, emojis, and emoticons in the text. Our independent-sample $t$-tests confirmed the effectiveness of manipulations in actual texts: $t (443) = 21.46$, $p < .001$, Cohen’s $d = 2.01$ for tone informality, and $t (294) = -21.80$, $p < .001$, Cohen’s $d = -2.57$ for cultural tailoring. The ratio of colloquial words in the informal tone condition was 18.9\%, compared to 0.3\% in the formal tone condition. The cultural tailoring score was 3.76 in the tailored condition, compared to 0.86 in the generic condition.

To ensure the validity of GPT-4’s measures using human validation, three expert coders conducted binary coding of tone and cultural tailoring using 20 randomly sampled chat scripts (Krippendorf’s alpha: .95 for tone and .82 for cultural tailoring). The results showed strong Point-Biserial correlations between human judgments and GPT-4 measures, with 0.88 for tone and 0.73 for cultural tailoring (both $p < .001$), supporting the validity of GPT-4’s measures. Additionally, the GPT-4 measure of tone was validated against the LIWC dictionary, specifically the “Conversation” corpus (Boyd et al., 2022), using the full sample. The strong Pearson’s correlation of 0.73 ($p < .001$) between tone measured by GPT-4 and LIWC further supported the robustness of GPT-4’s tone measures.

The effectiveness of our manipulations was also supported using self-reported perceptions. For perceived tone, all participants indicated the chatbot’s communication style from three options: casual style, formal style, or do not remember. A significant majority (77.68\%) correctly recognized the assigned style, indicating effective tone manipulation: 
$\chi^2= 138.57$, $p < .001$, Cramér’s $V = 0.56$, suggesting a large effect size. For perceived cultural tailoring, participants evaluated how relevant the chatbot’s information was to the needs and interests of their community, specifically [Hispanic/Latino] or [Black/African American], on a 5-point scale (1 = very irrelevant, 5 = very relevant). Among Hispanic and Black participants, compared to those in a generic condition ($M = 3.38$, $SD = 1.17$, $n = 139$), participants assigned to interact with a culturally tailored chatbot reported perceiving the content as more relevant to their racial community ($M = 3.73$, $SD = 0.98$, $n = 154$). The difference was significant: $t(269) = -2.72$, $p = .007$, Cohen’s $d = -0.32$, suggesting a small effect size.


\subsection{Hypotheses and Research Questions}
For the full sample, the model-data fit was satisfactory (Figure 1): $\chi^2(92, N = 441) = 129.5$, $p = .006$, $\textrm{CFI} = 0.99$, $\textrm{SRMR} = 0.031$, $\textrm{RMSEA} = 0.030$, 90\% CI $[0.017, 0.042]$, $p = .998$. For the sub-sample, the model also had a good fit with the data (Figure 2): $\chi^2(92, N = 293) = 144.5$, $p = .001$, $\textrm{CFI} = 0.97$, $\textrm{SRMR} = 0.045$, $\textrm{RMSEA} = 0.044$, 90\% CI $[0.030, 0.058]$, $p = .752$. As both models yielded similar results, our report of findings focused on the full-sample model but relied on both models for specific hypotheses regarding cultural tailoring. 
 
\begin{figure}[!t]
\centering
\includegraphics[width=1.0\linewidth]{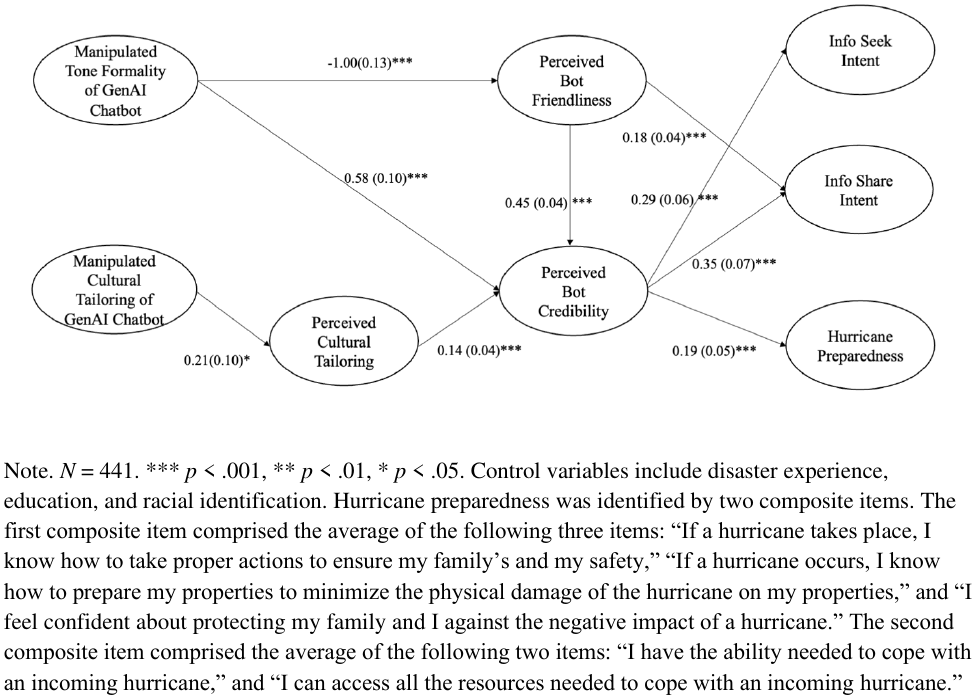}
\caption{Full Sample Results from SEM.}
\end{figure}

H1 predicted that the conversational tone of the GenAI chatbot significantly predicted diverse residents’ chatbot perceptions, including perceived friendliness and credibility, and H2 predicted that the chatbot’s perceived friendliness was positively associated with perceived credibility. Our results showed that the manipulated tone formality negatively predicted the perceived chatbot friendliness ($b = -1.00$, $SE = 0.13$, $p < .001$) and positively predicted the perceived bot credibility ($b = 0.58$, $SE = 0.10$, $p < .001$), supporting H1. Perceived friendliness of the chatbot was positively associated with perceived credibility: $b = 0.45$, $SE = 0.04$, $p < .001$, confirming H2. 

H3 hypothesized that the perceived friendliness of chatbots positively related to disaster preparedness outcomes including information-seeking intention, sharing intention, and hurricane preparedness, and H4 hypothesized that the perceived credibility of chatbots positively related to these disaster preparedness outcomes. Perceived chatbot friendliness positively related to the intent to share information with family and friends ($b = 0.18$, $SE = 0.04$, $p < .001$), but not information seeking or hurricane preparedness. H3 was partially supported. Additionally, perceived chatbot credibility positively related to information-seeking intention ($b = 0.29$, $SE = 0.06$, $p < .001$), sharing intention ($b = 0.35$, $SE = 0.07$, $p < .001$), and hurricane preparedness ($b = 0.19$, $SE = 0.05$, $p < .001$). H4 was fully supported. 

RQ2 asked how cultural tailoring of the interaction with AI chatbot affected diverse communities’ chatbot perceptions (perceived friendliness and credibility). Our results showed that manipulated cultural tailoring affected perceived cultural tailoring ($b = 0.21$, $SE = 0.10$, $p = .040$), which in turn positively affected perceived chatbot credibility ($b = 0.14$, $SE = 0.04$, $p < .001$). To validate the effect of cultural tailoring on chatbot perceptions, we also relied on the Hispanic and Black subsample. Figure 2 shows that cultural tailoring perceived by the Hispanic and Black participants significantly predicted perceived chatbot friendliness ($b = 0.33$, $SE = 0.13$, $p = .008$), which subsequently related to perceived bot credibility ($b = 0.64$, $SE = 0.13$, $p < .001$). Taken together, the findings suggested that perceived cultural tailoring predicted both perceptions differently for different racial communities. 

\begin{figure}[!t]
\centering
\includegraphics[width=1.0\linewidth]{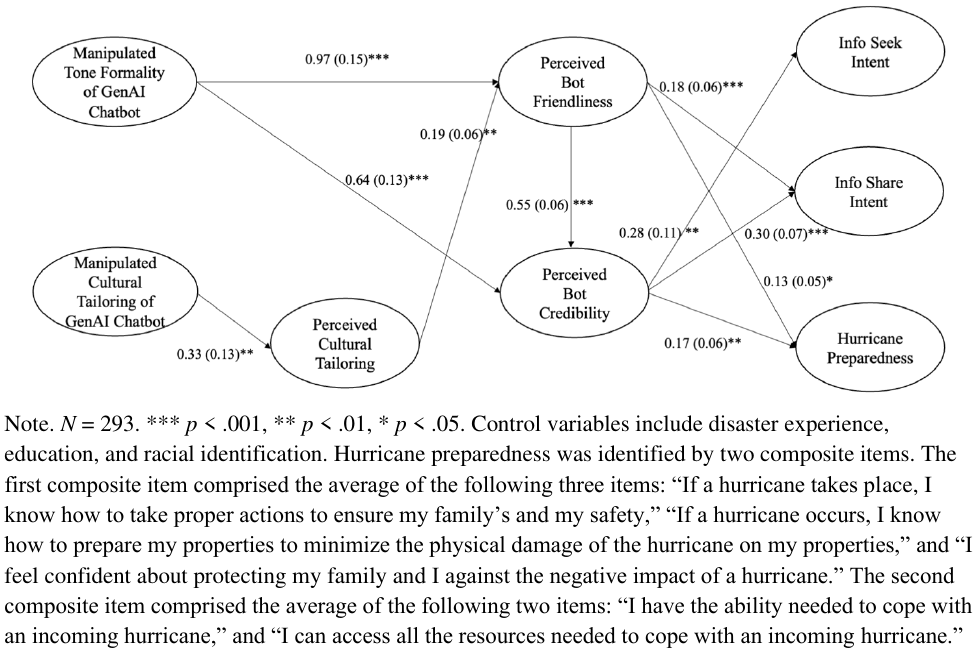}
\caption{The Hispanic and Black Sample Results from SEM.}
\end{figure}

A series of indirect effects were tested to answer RQ1 and RQ3. For RQ1, there was a significant indirect effect between the manipulated tone formality of the chatbot and information-sharing intent ($b = -0.18$, $SE = 0.048$, 95\% CI $[-0.303, -0.080]$) through bot friendliness. There was also a significant indirect effect from the manipulated tone formality to information sharing intent through bot credibility: $b = 0.17$, $SE = 0.04$, 95\% CI $[0.081, 0.287]$. For RQ3, there were indirect effects from perceived cultural tailoring to information-seeking intent ($b = 0.019$, $SE = 0.01$, 95\% CI $[0.007, 0.044]$), information-sharing intent ($b = 0.033$, $SE = 0.01$, 95\% CI $[0.010,0.073]$), and preparedness ($b = 0.022$, $SE = 0.01$, 95\% CI $[0.007,0.047]$) through bot credibility.

\section{Discussion}
Building upon the CASA paradigm and the literature on disaster vulnerability and cultural tailoring, this study designed GenAI chatbots in the context of disaster preparedness, testing the roles of tone informality and cultural tailoring in organization-public disaster communication. Leveraging OpenAI’s GPT-4 API that offers a scalable and flexible framework for programming-based customization of chatbots, our results from an online experiment with diverse communities show that GPT-4 chatbots varying in tone and cultural tailoring can significantly affect the perceived friendliness and credibility of chatbots, which are subsequently related to hurricane preparedness outcomes, including information seeking intent, information sharing intent, and preparedness. These results are discussed in detail as follows.

First, the SEM results show that the tone formality of GenAI chatbots was positively related to perceived bot credibility and negatively associated with perceived bot friendliness. This suggests the complex role of chatbot tone in disaster communication between an emergency management agency and diverse communities. On the one hand, a chatbot’s informal tone, as an important indicator of conversational human voice, increases perceived chatbot friendliness. On the other hand, a chatbot’s formal tone directly enhanced perceived source credibility, likely increasing information trustworthiness. Despite the inconsistency, our findings suggest that there was a positive net effect between tone formality and bot perceptions on hurricane preparedness outcomes. This thus suggests that from the standpoint of disaster management agencies, it is advised to predominantly use a formal tone while enhancing chatbot humanness through other appropriate elements, such as dialogic communication or humor. A contextually appropriate conversational tone can enhance perceived bot humanness without lowering trust or causing a backfire effect. This can potentially lead to higher information engagement in government–public disaster communication.

The open-ended responses from survey respondents further provided insights into the mixed results. When asked to reflect on the experience of interacting with the GenAI chatbots, while most participants approved the chatbot for providing useful and credible information, a few who were assigned to the informal tone condition indicated that the tone might be too casual to match the agency’s authority status. The perceived incongruency may elicit negative feelings toward the agency and the information obtained. Meanwhile, certain participants expressed favorable attitudes toward informal language cues, such as the use of emojis or the causal ways of greeting. The inconclusive findings thus invite future research to explore contextually and culturally appropriate ways to implement conversational human voice in the context of disasters.

Additionally, different GenAI chatbot perceptions facilitated disaster preparedness outcomes in distinct ways. Specifically, perceived bot friendliness was only positively associated with disaster information sharing intention. In contrast, perceived bot credibility consistently increased all three forms of disaster preparedness outcomes. Findings reaffirm the importance of source credibility and trust, well established in the HCI literature (e.g., Johnson \& Grayson, 2005). In disaster preparedness, the more credible individuals perceive the chatbots, the more likely they are motivated to take the next step of action to acquire and share information. Such credibility perception also translates into a higher level of disaster preparedness, which is consistent with existing disaster response literature in that trust and credibility perceptions of emergency management authorities can feed into greater disaster preparedness (e.g., Wachinger et al., 2013). In sum, our results suggest that enhancing the humanness of GenAI chatbots can improve public chatbot perceptions and engagement in the context of disaster preparedness, contributing to a community’s disaster resilience and preparedness.

Our results further underscore the potential of culturally tailored GenAI chatbots in disaster communication for multiethnic communities. Despite potential randomness in texts generated by GenAI chatbots compared to rule-based chatbots, our cultural tailoring manipulation successfully achieved both the actual degree of cultural tailoring in chat scripts and the perceived degree of cultural tailoring among Hispanic and Black participants. The robustness of our manipulation check indicates that cultural tailoring could be a promising strategy for enhancing GenAI chatbot communication with diverse cultural groups. Yet, the effect size of cultural tailoring was small, suggesting a discrepancy between the effectiveness of the intended and perceived cultural tailoring. This may be attributed to the use of similar system prompts across cultural groups: the lack of nuanced cultural tailoring might have diminished its effectiveness by unintentionally triggering stereotypes. Another reason could be our emphasis on surface-level tailoring more than deep-level tailoring, even though both were employed. Meta-analysis shows deep-level tailoring is typically more effective (Huang \& Shen, 2016). To increase the effectiveness of culturally tailored chatbots in disaster communication, future research could develop distinct prompts for specific cultural groups, balancing generic disaster information with culturally relevant values and beliefs. For Black communities, it might be essential to incorporate church and faith-based messaging, emphasize community solidarity, and acknowledge historical contexts. For Hispanic/Latino communities, cultural tailoring could focus on family values and promote collectivism.

Last, perceived cultural tailoring in the chatbot’s information enhanced its credibility, with slight variations across racial groups. Among Hispanic and Black participants, perceived cultural tailoring also improved perceptions of chatbot friendliness, which further reinforced credibility. Credible chatbots, in turn, contributed to higher information-seeking and sharing intentions, as well as better preparedness outcomes. The positive impact of cultural tailoring aligns with prior research on the significance of culture in health communication (Huang \& Shen, 2016) by highlighting the necessity of culturally appropriate disaster communication technologies. It also provides the first empirical evidence supporting its beneficial impact in the context of human-AI communication in disasters.

\subsection{Theoretical Implications}
Our study offers significant theoretical implications by extending the CASA paradigm (Nass et al., 1994) to the contexts of disaster communication and diverse cultural groups.

By establishing GenAI chatbots as information provision agents, this study investigates the roles of anthropomorphism through tone formality and cultural tailoring—unique dimensions of human-AI interactions in disasters. The integration of GenAI chatbots into a communicative approach to disaster management (e.g., Heath et al., 2009) offers a novel contribution to understanding how AI technologies can facilitate organization-public disaster communication, particularly in the under-investigated area of disaster preparedness. Specifically, our finding that the formal tone of GenAI chatbots increased perceived credibility but reduced friendliness in organizational disaster communication underscores the multifaceted nature of anthropomorphism in chatbot interactions during disasters. While disaster communication literature emphasizes the importance of formal, authoritative tones to enhance credibility (Zhao \& Tsang, 2021), HCI research highlights the value of chatbot friendliness in fostering user engagement and satisfaction (Jin \& Eastin, 2022). This points to the need for a more nuanced understanding when conceptualizing and designing the anthropomorphism of chatbots in applied communication scenarios such as disasters.

Traditionally, anthropomorphism in technology involves attributing human-like characteristics to non-human entities, such as giving chatbots human names or avatars (Go \& Sundar, 2019). However, the impact of cultural tailoring in improving perceived chatbot credibility suggests that users perceived our AI agents not only as social actors but also as culturally relevant ones. This implies that anthropomorphism could go beyond human-like features to include how these features resonate with users’ cultural backgrounds. Cultural tailoring represents a deeper dimension of chatbot anthropomorphism by aligning communication with specific cultural norms, values, and expectations—surpassing surface-level adjustments like using human names. This deeper form of anthropomorphism may foster stronger engagement between users and AI systems, potentially enhancing perceptions of machine intelligence and agency (Sundar, 2020). Future research could investigate whether incorporating various cultural components into chatbots strengthens perceived machine agency and how these factors influence strategic communication outcomes.

Furthermore, our study highlights the promising use of GenAI chatbots in experimental designs. Our results support that theoretical constructs such as cultural tailoring can be validly manipulated through GenAI chatbots via prompt engineering, as the robustness of our manipulation was confirmed by both textual analysis and user self-reports. To enhance the reproducibility and transparency of our research, we have made the complete source code for developing our chatbots openly available. By providing this open access, we encourage more sophisticated experiments in AI-mediated communication, exploring contexts beyond disaster communication to better understand the broader applicability and limitations of GenAI chatbots.

\subsection{Practical Implications}
This study also provides practical implications for disaster management agencies. First, the findings reiterate the importance of providing culturally tailored information for diverse community members. This can be especially meaningful in cultivating or restoring trust with a specific cultural community such as the African American community, which has long exhibited distrust of government and authorities due to many sociohistorical factors (Best et al., 2021). Disaster management agencies may also consider customizing formal versus informal communication styles when interacting with diverse cultural groups based on their needs, values, and traditions. While our study focuses on the disaster preparedness phase, suggesting that a formal tone may be more effective for delivering critical information and enhancing chatbot credibility, communication in other disaster stages may benefit from a more informal, friendly communication style, such as providing social support during or after disasters.

\subsection{Limitations and Future Directions}
This study has several limitations. First, our GPT-4 GenAI chatbot showed proficiency in disaster preparedness and could provide some local information. However, the absence of specific localized data, such as the updated nearby shelters, curtailed our ability to fine-tune the chatbot, sometimes resulting in generalized advice (e.g., urging users to visit official websites). One future direction involves augmenting GPT-4 with localized data through collaboration with agencies. We also acknowledge the challenge of aligning chatbot cultural tailoring with user perceptions for different cultural groups, which may reduce the effect size of the intervention. Future research should prioritize the creation of culturally specific prompts tailored to local communities (e.g., Haitian Americans). This approach could help bridge the gap between intended and perceived cultural alignment, ultimately enhancing disaster preparedness outcomes. In addition, our results should be generalized with caution, as they are based on a specific type of GenAI model (i.e., GPT 4). They may not be directly applicable to other generative AI models with different modalities, architectures, or training data. This highlights opportunities for future research to investigate the impact of various AI models in different communication contexts.

Another limitation of this study is the absence of direct measures for AI-related control variables such as AI chatbot use experience and AI literacy, which may influence users’ engagement with our GenAI chatbots. To mitigate this, we included educational level as a proxy control variable; however, future research should incorporate specific AI-related measures to better account for these potential confounding factors. Lastly, while this study primarily examined how the anthropomorphism of GenAI chatbots predicts disaster preparedness outcomes, future research could further explore information processing mechanisms. For instance, investigating how tone formality and cultural tailoring trigger specific machine heuristics and influence user engagement through the HAII-TIME model (Sundar, 2020) may provide deeper insights into the trust-building process.

\section{Conclusion}
This study enriches the CASA paradigm within the context of disaster communication and vulnerability and reveals the potential of GenAI Chatbots in experimental designs. Culturally tailored communication via GPT-4 chatbots to multiethnic communities can enhance chatbot perceptions and disaster preparedness. While humanizing chatbots through an informal tone can increase their perceived friendliness, it may also undermine their credibility and the effectiveness of disaster preparedness outcomes.

\section*{Supplementary Material}
Supplementary material is available online at Journal of Computer-Mediated Communication:
\url{https://academic.oup.com/jcmc/article/30/1/zmae022/7978204#supplementary-data}

\section*{Data Availability}
Data are available upon request to the corresponding author. For access to the data or inquiries regarding this article, please contact Xinyan Zhao at ezhao@unc.edu.

The source code of our chatbot and user instructions are available at: \url{https://bitbucket.org/leecwwong/ai_chatbot/}.\\

\textit{Conflicts of interest}: None declared.

\end{document}